\documentclass[journal]{IEEEtran}
\usepackage{graphicx,float,stfloats,amssymb,multirow,pifont,amsmath,scalerel,makecell}
\usepackage{bm,hyperref,xcolor,enumitem,booktabs,arydshln,footnote,url}
\usepackage[caption=false]{subfig}
\usepackage{color}
\usepackage{amsthm,amsmath,amssymb}
\usepackage{mathrsfs}
\usepackage{threeparttable}
\usepackage[subfigure,titles]{tocloft}
\usepackage[numbers,sort&compress]{natbib}
\hypersetup{pdftitle={Towards Text-Only Remote Sensing Scene Classification via Semantic Proxy},
    pdfauthor={Xianping Ma, Kunlin Wu, Xiaofeng Liu, Xiaokang Zhang, Yuanxin Ye},
	colorlinks=true,
	linkcolor=blue,
	anchorcolor=blue,
	citecolor=blue}

\graphicspath{{figures/}}

\begin{document}
\title{Towards Text-Only Remote Sensing Scene Classification via Semantic Proxy}
\author{Xianping~Ma,
		Kunlin~Wu,
		Xiaofeng~Liu,
		Xiaokang~Zhang,
		Yuanxin~Ye$^{\ast}$
		\thanks{Xianping~Ma and Yuanxin~Ye are with the Faculty of Geosciences and Engineering, Southwest Jiaotong University, Chengdu 611756, China.}
		\thanks{Kunlin~Wu and Xiaofeng~Liu are with the Society Hub, The Hong Kong University of Science and Technology (Guangzhou), China, Guangzhou 511453, China.}
		\thanks{Xiaokang~Zhang is with the School of Artificial Intelligence, Wuhan University, Wuhan 430072, China.}
}

\maketitle
\begin{abstract}
Remote sensing scene classification (RSSC) is traditionally formulated as a vision-centric problem, where inference relies on rich visual representations extracted from images. In this work, we revisit this assumption by introducing \emph{text-only RSSC}, a new problem setting that investigates whether remote sensing scenes can be recognized using only textual semantic representations. To enable this study, we propose SeProRS, which reformulates RSSC as a text-only classification task through semantic proxy generation. Specifically, a vision-language model (VLM) is first employed to convert remote sensing images into textual descriptions, termed semantic proxies, which are then encoded by a text embedding model for scene classification. This decoupled pipeline provides a unified and reproducible benchmark for evaluating text-only scene understanding. Furthermore, we investigate representative prompting strategies and proxy generators, and analyze their impact on classification performance, text embedding behavior, cross-generator transferability, and efficiency. Extensive experiments on three well-known RSSC datasets show that semantic proxies can preserve substantial discriminative information, while also revealing inherent limitations of purely text-based scene understanding. This paper does not aim to replace image-based RSSC, but instead establishes a new research direction by quantifying the feasibility and boundaries of semantic proxy-based scene recognition. The proposed benchmark and analysis provide a foundation for future research on efficient and modality-reduced remote sensing scene understanding. The source code and datasets will be accessible at \href{https://github.com/sstary/SSRS}{https://github.com/sstary/SSRS}.
\end{abstract}


\IEEEpeerreviewmaketitle

\section{Introduction}
Remote sensing scene classification (RSSC) aims to assign semantic land-use or land-cover labels to aerial or satellite image patches, and has long served as a fundamental task in remote sensing image understanding. Representative benchmarks, such as UCM \cite{yang2010ucmerced} and AID \cite{xia2017aid}, have promoted early progress by providing diverse scene categories and high-resolution image samples. Later datasets, such as NWPU-RESISC45 \cite{cheng2017nwpu}, further increased the number of categories and image variations, making RSSC a more challenging benchmark for evaluating scene-level visual representation learning \cite{cheng2020rsscDeep}. Driven by these benchmarks, RSSC has evolved from handcrafted descriptors and CNN-based models to Transformer-based architectures that better capture long-range spatial dependencies and contextual scene patterns \cite{bazi2021vision}. Despite continuous architectural advances, most existing RSSC methods share the same basic assumption: the target scene must be represented by visual features extracted from the input image.

\begin{figure}[t]
\centering
{\includegraphics[width=\linewidth]{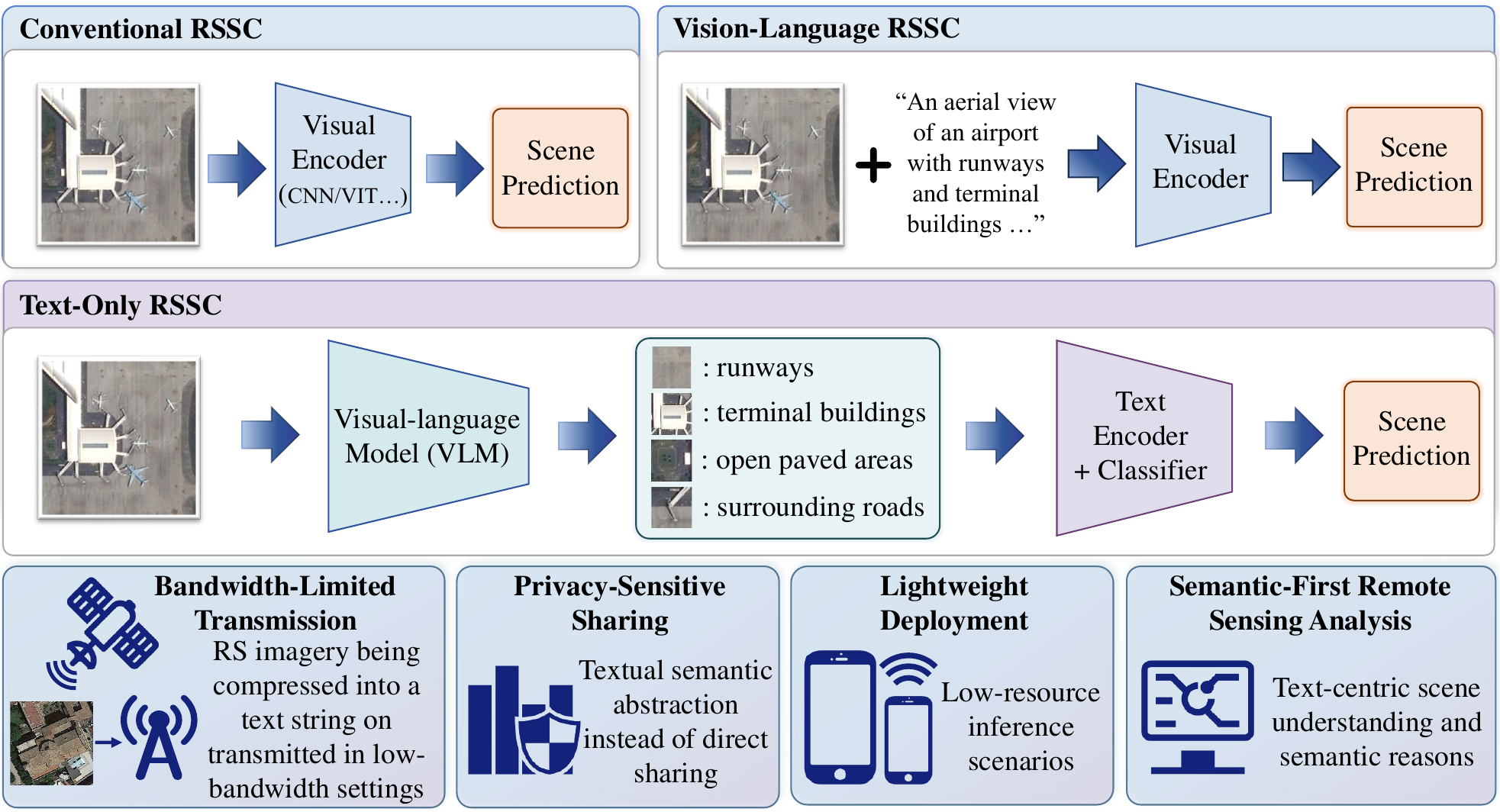}}
\caption{Comparison of three RSSC paradigms. (a) Conventional RSSC relies on image-derived visual features for scene classification, while (b) vision-language RSSC uses textual cues to assist visual recognition. (c) The proposed text-only RSSC instead represents remote sensing scenes with textual semantic proxies and predicts scene categories from text representations alone. Representative application scenarios are included to illustrate the practical relevance of this setting.}
\label{fig_tasks}
\end{figure}
This vision-centric paradigm has been further strengthened by the emergence of foundation models in remote sensing. Large-scale visual pretraining has enabled remote sensing models to learn transferable representations from massive unlabeled or weakly labeled imagery, improving their adaptability across sensors, resolutions, and downstream tasks \cite{cong2022satmae,reed2023scaleMAE}. Domain-specific foundation models, such as RingMo \cite{sun2023ringmo} and DOFA \cite{xiong2024dofa}, further demonstrate the potential of large-scale pretraining for capturing generic remote sensing visual patterns. Recent surveys also indicate that foundation models are becoming a central paradigm in remote sensing image interpretation, covering visual, multimodal, and promptable model families \cite{lu2025vision,huo2025rsfmSurvey}. However, these models still primarily improve the quality of visual representations, while the final prediction remains dependent on image pixels or image-derived features.

Recently, vision-language learning has introduced textual semantics into remote sensing interpretation. RS-CLIP \cite{li2023rsclip} and RemoteCLIP \cite{liu2024remoteclip} adapt contrastive vision-language learning to remote sensing and show promising performance in zero-shot scene classification and image-text retrieval. Large-scale image-text datasets and aligned VLMs, such as RS5M/GeoRSCLIP \cite{zhang2024rs5m} further expand the semantic supervision available for remote sensing representation learning. More recent studies explore prompt learning and transductive inference to improve zero-shot RSSC under vision-language settings \cite{el2025enhancing,dimitrovski2025few}. Meanwhile, remote sensing multimodal large language models, such as RS-LLaVA \cite{bazi2024rsllava} and SkyEyeGPT \cite{zhan2025skyeyegpt}, extend vision-language interpretation toward captioning, visual question answering, grounding, and instruction-following analysis. These developments show that language can provide rich semantic cues for remote sensing scenes, but they still generally use text to assist visual recognition rather than to replace visual inputs.

This observation raises an underexplored question: \emph{Can textual semantic representations serve as partial substitutes for visual inputs in remote sensing scene understanding?} As a first step, this paper instantiates this question on RSSC and examines whether scene categories can be inferred from textual semantic proxies alone. Unlike conventional vision-language RSSC, where text typically serves as a class prompt, auxiliary description, or alignment target while visual features remain indispensable \cite{li2023rsclip,el2025enhancing}, our setting treats text as the only classification input. For RSSC, this question is particularly meaningful because scene categories are often defined not only by object presence, but also by object co-occurrence, spatial layout, density, scale, and functional context. For example, airport, commercial area, dense residential area, and industrial area can share common objects such as buildings and roads, but differ substantially in spatial organization and land-use semantics \cite{xia2017aid,cheng2017nwpu}. Therefore, text-only RSSC requires more than class-name prompting or overly generic descriptions. The textual representation should preserve task-relevant visual evidence as much as possible, enabling downstream classification to infer scene categories from sufficient semantic cues. Such semantic proxies provide compact and interpretable scene-level information, making this paradigm potentially useful when direct image access, transmission, or deployment is constrained. Fig.~\ref{fig_tasks} compares conventional RSSC, vision-language RSSC, and the proposed text-only RSSC setting, while also highlighting several representative application scenarios.

Building upon this motivation, we take a first step toward this unexplored direction by introducing text-only RSSC, a new problem setting that investigates whether remote sensing scenes can be recognized from textual semantic representations alone. Rather than extending existing vision-language RSSC methods that still rely on visual inputs for final prediction, this work reformulates RSSC as a text-only semantic reasoning problem, where the visual scene is first converted into a textual semantic proxy and the final prediction is made solely from text representations. To establish a reproducible baseline for this new task, we propose SeProRS, a foundation model-driven semantic proxy framework consisting of two key stages: semantic proxy generation and text-based scene classification. Specifically, a VLM is employed to generate scene-level textual proxies from remote sensing images, while a text embedding model is used to encode these proxies into discriminative semantic representations for classification. This pipeline enables the feasibility of text-only RSSC to be systematically examined under a unified framework. Furthermore, we investigate two representative prompting strategies for semantic proxy generation, aiming to reveal how different forms of textual descriptions affect the behavior of text-only RSSC. Through this work, we do not claim to replace image-based RSSC, but instead establish a new research direction and provide a foundational benchmark for understanding how far textual semantic proxies can support RSSC when classification relies only on textual information. The main contributions of this work are summarized as follows:
\begin{itemize}[leftmargin=*]
\item We introduce text-only remote sensing scene classification, a new RSSC setting that explores scene recognition using textual semantic representations as the only classification input;
\item We propose SeProRS, a foundation model-driven baseline that performs text-only RSSC through VLM-based semantic proxy generation and text embedding-based classification;
\item We establish a systematic benchmark for text-only RSSC using three standard datasets, three representative VLMs, and two typical prompting strategies, revealing the feasibility and limitations of semantic proxy-based scene recognition.
\end{itemize}
To facilitate reproducible research and future comparison, we will release the generated semantic proxies on three standard RSSC datasets, establishing a reusable text-only resource for semantic proxy-based remote sensing recognition.

\section{Related Work}
\subsection{Remote Sensing Scene Classification}
RSSC methods mainly focus on learning discriminative visual representations from remote sensing images. Since scene categories are usually determined by complex combinations of objects, textures, spatial layouts, and contextual patterns, effective RSSC requires representations that can capture both local visual details and global scene structures. Early deep learning methods addressed this problem by transferring or fine-tuning CNNs pretrained on natural images, showing that hierarchical visual features are effective for remote sensing scene recognition \cite{hu2015transferring}. Subsequent studies further improved visual representation learning to enhance the separability of visually similar scene categories \cite{cheng2020rsscDeep}. Therefore, the mainstream development of RSSC has been centered on strengthening image-derived visual features.

CNN-based methods first dominated the visual representation-driven paradigm in RSSC. By learning hierarchical features from image pixels, deep convolutional models provide stronger scene representations than handcrafted descriptors and improve the recognition of complex land-use categories \cite{nogueira2017towards}. Later studies further enhanced CNN representations through multi-scale feature extraction, attention mechanisms, and contextual modeling, aiming to capture local texture, object composition, and category-specific spatial patterns more effectively \cite{wang2022transferring, yang2026msalnet}. Transformer-based methods then introduced global self-attention into RSSC, enabling long-range dependency modeling and more flexible scene-structure representation \cite{bazi2021vision, tong2025SFormer}. Hybrid CNN-Transformer designs further combine local inductive bias with global modeling ability, reflecting a continued effort to improve image-derived visual representations for scene classification \cite{sivasubramanian2024transformer}.

Foundation models have pushed this visual paradigm further by transferring large-scale pretrained representations to remote sensing tasks \cite{li2024vision,lu2025vision}. More recent efforts, including DINOv3 with satellite pretraining \cite{simeoni2025dinov3}, SkySense++ \cite{wu2025semantic}, and AlphaEarth Foundations \cite{brown2025alphaearth}, further demonstrate the trend toward larger-scale visual or geospatial representation learning for Earth observation. These developments significantly strengthen the visual backbone of RSSC. However, existing RSSC methods remain fundamentally vision-centric. The role of non-visual semantic representations as independent carriers for scene classification has not been sufficiently explored.

\subsection{Vision-Language Learning in Remote Sensing}
Vision-language learning has become an important direction in remote sensing interpretation. Early remote sensing image captioning datasets, such as RSICD \cite{lu2018rsicd} and RSITMD \cite{yuan2022rsitmd}, demonstrate that aerial images can be associated with natural language descriptions containing objects, attributes, and contextual relations. More recent large-scale image-text datasets further expand remote sensing vision-language learning by collecting or generating large amounts of paired satellite images and textual descriptions \cite{yuan2025chatearthnet}. These datasets provide a foundation for learning cross-modal alignment between remote sensing imagery and language.

These image-text resources have supported the rapid evolution of remote sensing vision-language learning \cite{weng2025vlmrs,tao2025advancements}. Recently, LRSCLIP \cite{chen2025lrsclip} extends long-text vision-language modeling to remote sensing and improves cross-modal retrieval and zero-shot recognition under more descriptive textual supervision. LHRS-Bot-Nova \cite{li2025lhrsbotnova} enhances vision-language alignment through improved visual encoding, visual compression, and remote sensing-oriented instruction data. These efforts further explore reliable instruction generation and hierarchical task modeling for remote sensing VLMs, such as DDFAV \cite{li2025ddfav} and GeoRSMLLM \cite{zhang2025georsmllm}. 

Although these studies demonstrate the growing importance of language in remote sensing, their primary goal is still to enhance visual interpretation. Text is typically used as supervision, prompt, instruction, or response, while visual encoders and image features remain indispensable. Therefore, existing remote sensing vision-language studies have not systematically examined whether textual semantic representations, once generated from remote sensing images, can serve as independent inputs for remote sensing scene understanding.

\begin{figure}[t]
\centering
{\includegraphics[width=\linewidth]{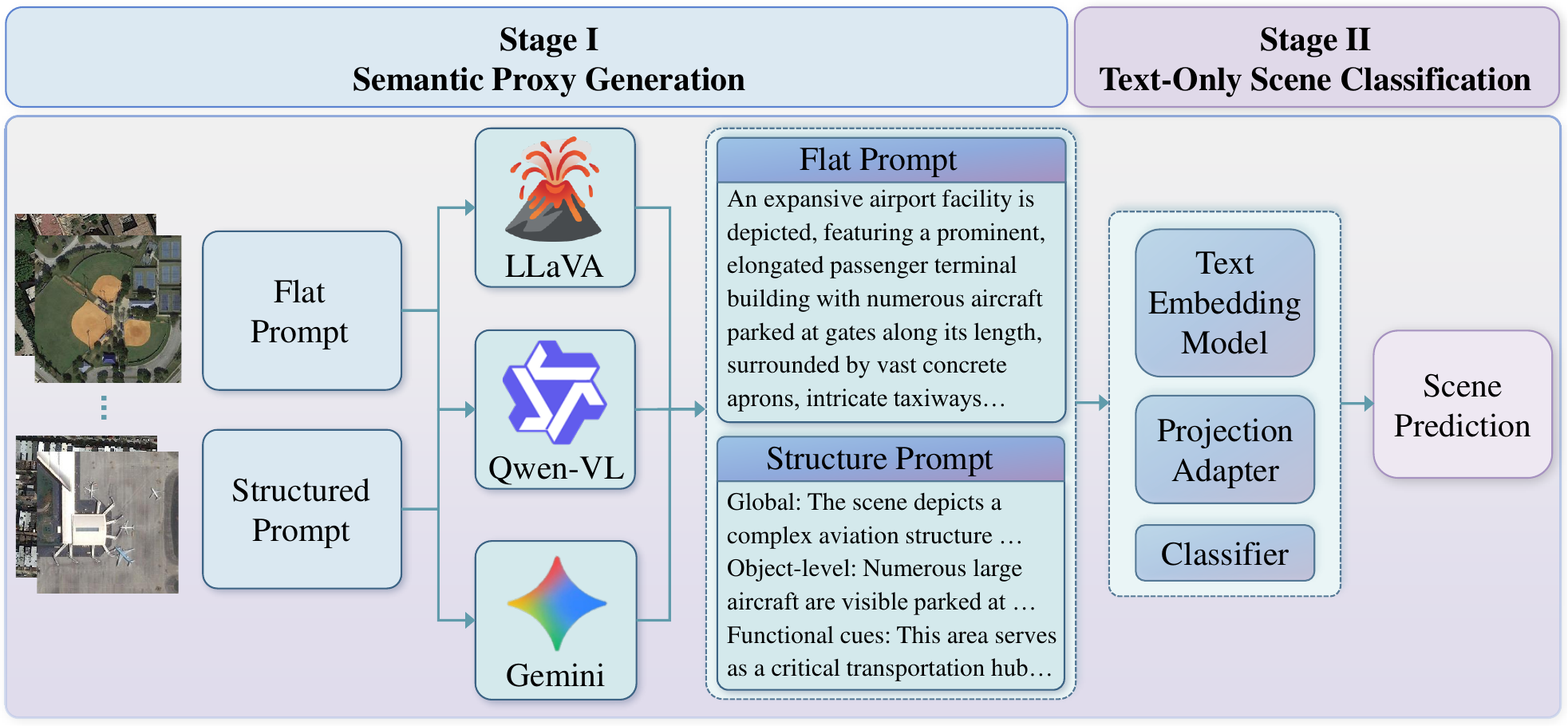}}
\caption{Overall framework of SeProRS. Remote sensing images are first converted into textual semantic proxies by VLMs under task-agnostic descriptive prompts, and the downstream model performs scene recognition using only the generated proxy. The generated proxies are stored and reused for all text-only training and testing, which decouples semantic abstraction from classifier optimization.}
\label{fig_framework}
\end{figure}
\section{Text-Only Remote Sensing Scene Classification}
\subsection{Problem Formulation}
Conventional RSSC is typically formulated as an image-based recognition problem. Given a remote sensing image $x \in \mathcal{X}$ and its corresponding scene label $y \in \mathcal{Y}$, where $\mathcal{Y}={1,\ldots,C}$ denotes a predefined set of $C$ scene categories, the goal is to learn a visual classifier:
\begin{equation}
f_v: \mathcal{X} \rightarrow \mathcal{Y},
\end{equation}
which predicts the scene category from image pixels or image-derived visual features. Under this formulation, inference depends on visual evidence extracted from $x$.

In contrast, this work considers \emph{text-only remote sensing scene classification}, which represents a fundamentally different setting. Instead of directly classifying an image, this setting assumes that each scene is represented by a textual semantic proxy $s \in \mathcal{S}$, where $\mathcal{S}$ denotes the space of natural-language scene descriptions. The classifier then learns a text-based mapping:
\begin{equation}
f_t: \mathcal{S} \rightarrow \mathcal{Y},
\end{equation}
and predicts the scene category using only $s$ as input. This setting does not imply that images are absent from the entire pipeline; rather, it specifies that the final prediction is made from textual representations without accessing image pixels or visual features.

More formally, text-only RSSC can be viewed as a two-step abstraction-and-recognition problem. A semantic abstraction function $\phi$ converts an image into a textual proxy:
\begin{equation}
s = \phi(x),
\end{equation}
and a text-based classifier $f_t$ predicts the label from this proxy:
\begin{equation}
\hat{y} = f_t(s) = f_t(\phi(x)).
\end{equation}
The key distinction from conventional RSSC lies in the information available to the classifier: while $f_v$ directly accesses visual evidence from $x$, $f_t$ only receives the textual abstraction $s$.

\begin{table*}[t]
	\centering
	\caption{Comparison of different textual representations related to RSSC.}\label{tab:text_representation} 
	\begin{tabular}{lccc} 
		\toprule 
		\textbf{Representation} & \textbf{Granularity} & \textbf{Image-specific} & \textbf{Main Function} \\
		\midrule Class label & Category-level & No & Defines the prediction target \\ 
		Class prompt & Category-level & No & Provides textual class prior \\ 
		Generic description & Instance-level & Yes & Describes visible scene content \\ 
		Semantic proxy & Instance-level & Yes & Preserves task-relevant evidence for text-only classification \\ 
		\bottomrule 
	\end{tabular} 
\end{table*}
\subsection{Semantic Proxy}
In the text-only RSSC setting, the semantic proxy serves as the intermediate representation that connects the original visual scene to text-based classification. It is generated for a specific remote sensing image and used as the classification input. Importantly, the term semantic proxy does not refer to a fixed linguistic format. It may be expressed as a free-form scene description or as a structured textual description, depending on the proxy-generation strategy. What makes it a proxy is its functional role: it carries scene-relevant semantic evidence in textual form and replaces direct visual inputs during classification. As summarized in Table~\ref{tab:text_representation}, the semantic proxy is therefore distinguished by both its granularity and its role in the task.

For example, rather than simply associating an image with the word ``airport'', a proxy may describe evidence such as runways, large paved areas, terminal-like buildings, and surrounding transportation structures. Such textual evidence provides a basis for predicting the scene category from semantic cues rather than from raw visual features. The quality of a semantic proxy is therefore determined by its usefulness for downstream scene recognition, not by whether it follows a specific textual style. A fluent caption-like description, a concise summary, or a structured semantic description can all serve as semantic proxies if they preserve discriminative scene information for classification.

\subsection{Scope and Implications}
Text-only RSSC should be interpreted as an initial exploration of semantic proxy-based remote sensing understanding, rather than as a direct replacement for image-based RSSC. In this work, RSSC is selected as a representative scene-level task because its prediction depends on high-level land-use semantics, object co-occurrence, spatial organization, and contextual cues. These properties make RSSC a suitable testbed for examining whether textual semantic representations can partially substitute visual inputs for scene-level recognition.

This setting also defines a different evaluation perspective from conventional visual recognition. Since the classifier only accesses textual proxies, its performance reflects the information-preserving capacity of semantic abstraction and the discriminability of text-based scene representations. Therefore, the results should not be interpreted as a direct upper bound of image-based RSSC, where richer visual evidence is available. Instead, they provide evidence about the feasibility and limitations of performing remote sensing recognition under a modality-reduced condition.

Beyond RSSC, this formulation points to a broader research direction. Many remote sensing applications involve semantic-level decisions, where compact textual representations may support tasks such as efficient transmission, privacy-sensitive sharing, and lightweight deployment. In this sense, text-only RSSC provides a concrete benchmark for studying how far semantic proxies can support modality-reduced remote sensing understanding.

\section{SeProRS: A Semantic Proxy Baseline}
\subsection{Overall Pipeline}
SeProRS is designed as a simple and reproducible baseline framework for text-only RSSC, rather than a task-specific architecture with complex modules. The purpose of this work is to establish a standard pipeline for studying semantic proxy-based recognition. Therefore, each component in SeProRS is intentionally kept simple, so that the resulting performance mainly reflects the effectiveness and limitations of textual semantic proxies, instead of the capacity of a specialized classifier. The overall framework of SeProRS is illustrated in Fig.~\ref{fig_framework}.

Given a remote sensing image $x_i$, SeProRS first generates a textual semantic proxy $s_i$ through a VLM. The proxy is then encoded into a text representation and used for scene prediction. The overall process can be written as:
\begin{equation}
    s_i = \Phi_{\mathrm{vlm}}(x_i, p),
\end{equation}
\begin{equation}
    \mathbf{z}_i = E_{\mathrm{text}}(s_i),
\end{equation}
\begin{equation}
    \hat{y}_i = g(\mathbf{z}_i),
\end{equation}
where $\Phi_{\mathrm{vlm}}$ denotes the VLM-based proxy generator, $p$ is the prompting instruction, $E_{\mathrm{text}}$ is a text embedding model, $\mathbf{z}_i$ is the text representation, and $g$ is a lightweight prediction function. In this pipeline, the image is used only for proxy generation. Once $s_i$ is obtained, both training and testing of the classifier are conducted entirely in the textual space without accessing image pixels or visual features.

This design distinguishes SeProRS from direct VLM-based classification. Instead of asking a VLM to directly output a scene label, SeProRS uses the VLM to produce an intermediate semantic proxy and then evaluates whether this proxy can support downstream classification. This separation makes the benchmark more interpretable, because the generated text can be inspected as the modality-reduced representation that mediates between the original image and the final prediction.

\subsection{VLM-based Semantic Proxy Generation}
The first stage of SeProRS converts each remote sensing image into a textual semantic proxy. Since the proxy serves as the only input for subsequent classification, the VLM is treated as a semantic abstraction module rather than as a classifier. Its role is to translate visual observations into natural language while preserving scene-relevant evidence useful for RSSC.

For a dataset $\mathcal{D}=\{(x_i,y_i)\}_{i=1}^{N}$, each image $x_i$ is processed by a VLM under a given prompt $p$. This produces a text-only dataset:
\begin{equation}
    \mathcal{D}_{t}^{(p)}=\{(s_i^{(p)},y_i)\}_{i=1}^{N},
\end{equation}
where $s_i^{(p)}=\Phi_{\mathrm{vlm}}(x_i,p)$ denotes the generated proxy. When multiple VLMs and prompting strategies are used, the corresponding proxy sets form different text-only views of the same RSSC dataset. This allows the text-only setting to be evaluated under different semantic abstraction behaviors without changing the downstream classifier.

In our protocol, proxy generation is deliberately label-free: no class name, ground-truth label, or candidate label list is provided to the VLM. Each proxy is generated once and stored before classifier training, so downstream experiments can be repeated without calling the VLM again. This protocol prevents the VLM from acting as a label selector and makes the generated text a reusable benchmark resource.

\subsection{Prompting Strategies}
To examine how proxy formulation affects text-only RSSC, SeProRS adopts two representative prompting strategies: \emph{Flat Prompt} and \emph{Structured Prompt}. These two strategies are not designed as sophisticated prompt engineering schemes, but as simple and reproducible ways to produce different forms of semantic proxies.

The Flat Prompt asks the VLM to describe the remote sensing scene in a concise natural-language paragraph. This strategy encourages a fluent scene-level description that integrates overall scene type, key objects, and functional context when possible. It resembles a caption-like proxy, but its role is different from conventional captioning: the generated text is directly used as the classification input rather than being evaluated for linguistic quality.

The Structured Prompt asks the VLM to describe the scene through multiple semantic aspects in a fixed order. In our implementation, the prompt guides the model to produce descriptions related to the global scene, object-level attributes, and functional cues. This strategy encourages the proxy to expose different types of scene evidence in a more organized form. However, the use of a structured prompt does not imply that structured text is necessarily superior. Instead, the comparison between Flat and Structured prompts is intended to reveal how different proxy formats influence the feasibility and behavior of text-only RSSC.

\subsection{Text-Only Scene Classification}
After semantic proxy generation, SeProRS performs scene classification using only textual representations. Each proxy $s_i$ is encoded by the text embedding model $E_{\mathrm{text}}$.

To avoid introducing unnecessary architectural complexity, we adopt a lightweight prediction module on top of the text embeddings. Before classification, $\mathbf{z}_i$ is passed through a small projection adapter consisting of layer normalization, a linear layer, GELU activation, and dropout. This adapter slightly adjusts the general-purpose text embedding space to the target RSSC label space without introducing a task-specific architecture. The adapted representation $\tilde{\mathbf{z}}_i$ is then fed into a linear classifier:
\begin{equation}
    \hat{\mathbf{q}}_i = \mathrm{softmax}(\mathbf{W}\tilde{\mathbf{z}}_i+\mathbf{b}),
\end{equation}
where $\mathbf{W}\in\mathbb{R}^{C\times h}$ and $\mathbf{b}\in\mathbb{R}^{C}$ are learnable parameters, $h$ is the projected feature dimension, and $C$ denotes the number of scene categories. The training objective is the standard cross-entropy loss.

The use of a simple text embedding model and lightweight classifier is deliberate. Since the focus of this work is the text-only setting and the semantic proxy representation, a simple classification network helps prevent the benchmark from being dominated by task-specific model design. Consequently, the resulting performance provides a clearer indication of how much discriminative scene information is preserved in the generated textual proxies and how effectively such information can support RSSC under a modality-reduced inference protocol.

\section{Experiments}
\subsection{Datasets and Settings}

\begin{figure}[t]
\centering
{\includegraphics[width=\linewidth]{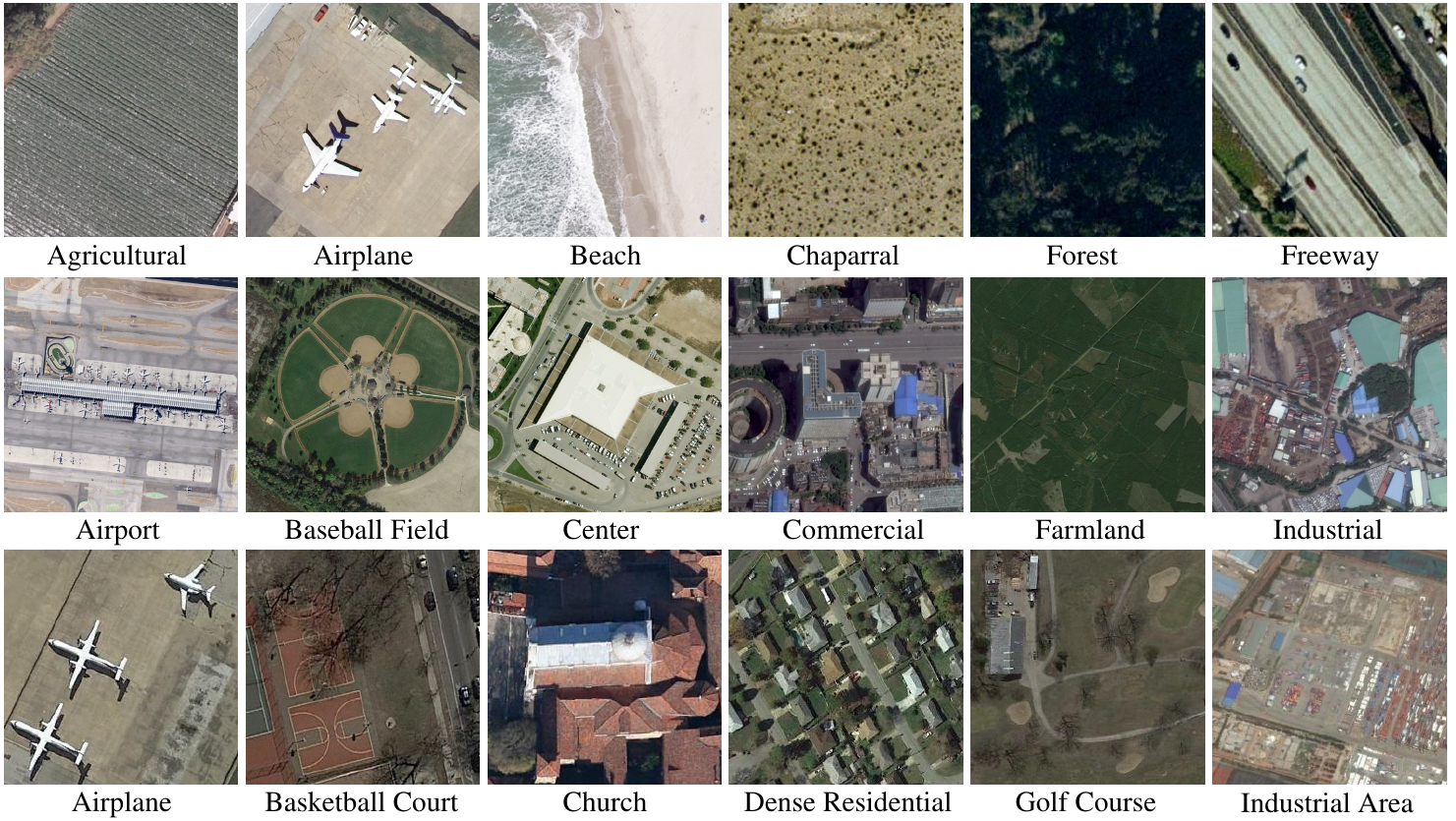}}
\caption{Representative samples from UCM, AID, and NWPU-RESISC45 datasets, with 6 samples selected from each.}
\label{fig_samples}
\end{figure}

\begin{table*}[t]
\centering
\caption{Summary of the RSSC datasets and experimental splits used in SeProRS.}
\label{tab:dataset_summary}
\begin{tabular}{lccccc}
\toprule
\textbf{Dataset} & \textbf{Classes} & \textbf{Total images} & \textbf{Spatial resolution} & \textbf{Training/Test ratio} & \textbf{Train/Test images} \\
\midrule
\multirow{2}{*}{UCM} & \multirow{2}{*}{21} & \multirow{2}{*}{2,100} & \multirow{2}{*}{0.3 m} & 50\% / 50\% & 1,050 / 1,050 \\
 & & & & 80\% / 20\% & 1,680 / 420 \\
\midrule
\multirow{2}{*}{AID} & \multirow{2}{*}{30} & \multirow{2}{*}{10,000} & \multirow{2}{*}{0.5-8 m} & 20\% / 80\% & 2,000 / 8,000 \\
 & & & & 50\% / 50\% & 5,000 / 5,000 \\
\midrule
\multirow{2}{*}{NWPU-RESISC45} & \multirow{2}{*}{45} & \multirow{2}{*}{31,500} & \multirow{2}{*}{0.2-30 m} & 10\% / 90\% & 3,150 / 28,350 \\
 & & & & 20\% / 80\% & 6,300 / 25,200 \\
\bottomrule
\end{tabular}
\end{table*}

SeProRS is evaluated on three widely used RSSC benchmarks, including UCM~\citep{yang2010ucmerced}, AID~\citep{xia2017aid} and NWPU-RESISC45~\citep{cheng2017nwpu}. These datasets cover different spatial resolutions, category scales, and scene complexities, providing a controlled basis for analyzing whether semantic-only representations preserve discriminative scene information. The detailed dataset statistics and training/testing splits are summarized in Table~\ref{tab:dataset_summary}, and representative samples are shown in Fig.~\ref{fig_samples}.

\textbf{UCM} contains 2,100 images from 21 land-use categories, with 100 images per category. Each image is of size $256 \times 256$ with a spatial resolution of 0.3\,m. In this work, UCM is included as an explicit benchmark under 50\%/50\% and 80\%/20\% splits, enabling a compact yet informative evaluation of semantic-only classification under relatively small-scale data settings.

\textbf{AID} contains 10,000 aerial scene images from 30 categories, with a spatial size of $600 \times 600$ and spatial resolutions ranging from 0.5\,m to 8\,m. Following common RSSC protocols, 20\% and 50\% of images are used for training, while the remaining samples are used for testing.

\textbf{NWPU-RESISC45} contains 31,500 images evenly distributed across 45 scene categories. Each image has a size of $256 \times 256$, and the spatial resolution varies from 0.2\,m to 30\,m. We adopt the 10\%/90\% and 20\%/80\% training/testing splits, which are widely used in RSSC evaluation and provide a challenging testbed for semantic scene discrimination.

\begin{table*}[t]
\centering
\caption{Visual-language models used for semantic proxy generation in SeProRS.}
\label{tab:semantic_generators}
\begin{tabular}{llllll}
\toprule
\textbf{ID} & \textbf{Generator} & \textbf{Access} & \textbf{Model Scale} & \textbf{Parameters} & \textbf{Role} \\
\midrule
G1 & LLaVA-OneVision-Qwen2-0.5B~\citep{li2024llava} & local & Small & 0.5B & lightweight open VLM \\
G2 & Qwen2.5-VL-3B~\citep{bai2025qwen25} & local & Medium & 3B & efficient open VLM \\
G3 & Gemini-2.5-Flash~\citep{comanici2025gemini} & API & Commercial & Not disclosed & closed-source strong VLM \\
\bottomrule
\end{tabular}
\end{table*}

As summarized in Table~\ref{tab:semantic_generators}, we consider one commercial VLM, Gemini-2.5-Flash, one efficient open VLM, Qwen2.5-VL-3B, and one lightweight LLaVA-OneVision model. Gemini-2.5-Flash is used as a strong API-based generator, whereas Qwen2.5-VL-3B and LLaVA-OneVision-Qwen2-0.5B are deployed locally to examine the effect of access mode, model scale, and architecture on semantic proxy quality. The dataset class names are used only for defining the supervised training labels of the downstream classifier; they are not included in the proxy-generation prompts.

\begin{table}[t]\small
\centering
\caption{Hyperparameters for semantic proxy generation.}
\label{tab:proxy_generation_hyperparams}
\begin{tabular}{lp{0.58\columnwidth}}
\toprule
\textbf{Item} & \textbf{Setting} \\
\midrule
Input & Remote sensing image and task-agnostic descriptive prompt. \\
Label control & No class name, ground-truth label, or candidate label list is provided to the VLM. \\
Generators & LLaVA-OneVision-Qwen2-0.5B, Qwen2.5-VL-3B, Gemini-2.5-Flash. \\
Prompt types & Flat Prompt and Structured Prompt. \\
Flat Prompt & Generates one concise paragraph describing scene-level semantics. \\
Structured Prompt & Generates global description, object-level attributes, and functional cues. \\
Storage & Semantic proxies are generated once and then reused for all classifier training and testing. \\
\bottomrule
\end{tabular}
\end{table}

\subsection{Implementation Details}
All semantic-only classifiers are trained under the same data splits as the image-based RSSC protocols. Unless otherwise specified, the text encoder is frozen and only the projection adapter and classification head are optimized. The maximum text length is fixed to 192 tokens, and generated proxies longer than this limit are truncated. Each experiment is repeated five times with independent seeds, and the mean and standard deviation of overall accuracy (OA) are reported.

The experiments are designed to answer three questions: 1) how well can semantic-only proxies support RSSC; 2) whether structured proxies are more effective than flat paragraph-style proxies; and 3) how sensitive the classification performance is to the VLM used for semantic proxy generation. The same text-only classifier and training protocol are used across all settings.

The semantic proxy generation settings are reported in Table~\ref{tab:proxy_generation_hyperparams}, which specifies the task-agnostic descriptive prompts and the exclusion of class labels or candidate label lists. Additional implementation settings are reported in Tables~\ref{tab:text_hyperparams} and~\ref{tab:visual_hyperparams} of Appendix~\ref{app:implementation}. Table~\ref{tab:text_hyperparams} summarizes the text-only classifier configuration, including the frozen text encoder, projection adapter, optimization settings, and repeated runs. Table~\ref{tab:visual_hyperparams} lists the training settings for the vision-based reference models to make the visual comparison reproducible. For both text-only and vision-based training, the base learning rate of $5\times10^{-4}$ is linearly scaled by the global batch size divided by 512, yielding an effective learning rate of $1.25\times10^{-4}$ at a global batch size of 128. In addition, Tables~\ref{tab:proxy_examples}-\ref{tab:proxy_examples_nwpu} in Appendix~\ref{app:proxies} provide representative proxies from the three datasets under both Flat and Structured prompts for all three VLMs, illustrating the actual textual inputs used for downstream classification.

\subsection{Main Results}
To provide a visual-performance reference, which can be regarded as an upper bound under full image access, comparisons are conducted against representative vision-based RSSC models, including HFAM~\citep{wan2023hfam}, CGINet~\citep{zhao2024co}, UPetu~\citep{dong2024upetu}, MSCN~\citep{ma2025multiscale}, UniRepLKNet~\citep{zhang2025scaling} and DINOv3-Adapter. In particular, DINOv3-Adapter applies the standard adapter tuning strategy \citep{wu2025medical} to the recent vision foundation model DINOv3 \citep{simeoni2025dinov3}, and therefore serves as a strong and up-to-date backbone baseline in our experiments.

Tables~\ref{tab:ucm_image_reference}-\ref{tab:nwpu_image_reference} report the main results on UCM, AID, and NWPU-RESISC45. The vision-based methods are included as full-image reference results, while SeProRS performs classification only from textual semantic proxies. Therefore, these two groups should not be interpreted as input-matched competitors. Instead, the comparison provides a reference for understanding how much scene-discriminative information can be retained after remote sensing images are abstracted into text.

\begin{table}[t]
\centering
\caption{Main results on the UCM dataset. Vision-based methods use full image inputs, while SeProRS performs classification from textual semantic proxies.}
\label{tab:ucm_image_reference}
\begin{tabular}{lcc}
\toprule
\textbf{Method} & \textbf{50\% training} & \textbf{80\% training} \\
\midrule
\multicolumn{3}{c}{\textbf{Vision-based Methods}} \\
\midrule
HFAM & $98.35 \pm 0.14$ & $99.43 \pm 0.13$ \\
CGINet & $99.14 \pm 0.16$ & $99.84 \pm 0.16$ \\
UPetu & $98.95 \pm 0.12$ & $99.05 \pm 0.28$ \\
MSCN & $99.26 \pm 0.07$ & $99.95 \pm 0.10$ \\
UniRepLKNet & $99.37 \pm 0.08$ & $99.24 \pm 0.10$\\
DINOv3-Adapter & $99.12 \pm 0.07$ & $99.52 \pm 0.05$ \\
\midrule
\multicolumn{3}{c}{\textbf{SeProRS (Text-based Methods)}} \\
\midrule
G1 + Structured & $88.42 \pm 0.30$ & $89.48 \pm 0.52$ \\
G1 + Flat & $90.27 \pm 0.14$ & $91.24 \pm 0.31$ \\
G2 + Structured & $77.31 \pm 0.21$ & $80.90 \pm 0.20$ \\
G2 + Flat & $81.41 \pm 0.36$ & $83.81 \pm 0.48$ \\
G3 + Structured & $93.60 \pm 0.14 $& $92.16 \pm 0.07$ \\
G3 + Flat & $96.19 \pm 0.13$ & $95.39 \pm 0.27$ \\
\bottomrule
\end{tabular}%
\end{table}

\begin{table}[t]
\centering
\caption{Main results on the AID dataset. Vision-based methods use full image inputs, while SeProRS performs classification from textual semantic proxies.}
\label{tab:aid_image_reference}
\begin{tabular}{lcc}
\toprule
\textbf{Method} & \textbf{20\% training} & \textbf{50\% training} \\
\midrule
\multicolumn{3}{c}{\textbf{Vision-based Methods}} \\
\midrule
HFAM & $94.71 \pm 0.06$ & $96.73 \pm 0.04$ \\
CGINet & $95.34 \pm 0.14$ & $97.10 \pm 0.24$ \\
UPetu & $95.33 \pm 0.10$ & $97.04 \pm 0.21$ \\
MSCN & $95.86 \pm 0.16$ & $97.46 \pm 0.12$ \\
UniRepLKNet & $96.46 \pm 0.07$ & $97.82 \pm 0.10$ \\
DINOv3-Adapter & $\mathbf{97.58 \pm 0.11}$ & $\mathbf{98.62 \pm 0.05}$ \\
\midrule
\multicolumn{3}{c}{\textbf{SeProRS (Text-based Methods)}} \\
\midrule
G1 + Structured & $86.85 \pm 0.09$ & $89.28 \pm 0.14$ \\
G1 + Flat & $90.44 \pm 0.15$ & $92.01 \pm 0.07$ \\
G2 + Structured & $83.88 \pm 0.04$ & $85.96 \pm 0.12$ \\
G2 + Flat & $87.28 \pm 0.10$ & $88.51 \pm 0.13$ \\
G3 + Structured & $93.46 \pm 0.11$ & $94.76 \pm 0.06$ \\
G3 + Flat & $93.95 \pm 0.13$ & $95.18 \pm 0.04$ \\
\bottomrule
\end{tabular}%
\end{table}

\begin{table}[t]
\centering
\caption{Main results on the NWPU-RESISC45 dataset. Vision-based methods use full image inputs, while SeProRS performs classification from textual semantic proxies.}
\label{tab:nwpu_image_reference}
\begin{tabular}{lcc}
\toprule
\textbf{Method} & \textbf{10\% training} & \textbf{20\% training} \\
\midrule
\multicolumn{3}{c}{\textbf{Vision-based Methods}} \\
\midrule
HFAM & $90.81 \pm 0.11$ & $92.93 \pm 0.17$ \\
CGINet & $92.28 \pm 0.17$ & $94.38 \pm 0.13$ \\
UPetu & $92.20 \pm 0.07$ & $93.70 \pm 0.14$ \\
MSCN & $92.64 \pm 0.09$ & $94.59 \pm 0.11$ \\
UniRepLKNet & $93.61 \pm 0.06$ & $95.26 \pm 0.11$ \\
DINOv3-Adapter & $\mathbf{94.61 \pm 0.10}$ & $\mathbf{95.89 \pm 0.04}$ \\
\midrule
\multicolumn{3}{c}{\textbf{SeProRS (Text-based Methods)}} \\
\midrule
G1 + Structured & $76.10 \pm 0.10$ & $78.15 \pm 0.10$ \\
G1 + Flat & $83.79 \pm 0.06$ & $85.51 \pm 0.05$ \\
G2 + Structured & $67.66 \pm 0.11$ & $69.67 \pm 0.08$ \\
G2 + Flat & $73.39 \pm 0.16$ & $75.26 \pm 0.13$ \\
G3 + Structured & $87.64 \pm 0.06$ & $88.64 \pm 0.07$ \\
G3 + Flat & $89.13 \pm 0.08$ & $90.08 \pm 0.06$ \\
\bottomrule
\end{tabular}%
\end{table}

The results demonstrate that text-only RSSC achieves non-trivial performance across all three datasets, showing that semantic proxies preserve substantial scene-level information for classification. With the strongest proxy generator and the Flat Prompt, SeProRS reaches $96.19\%$ and $95.39\%$ on UCM, $93.95\%$ and $95.18\%$ on AID, and $89.13\%$ and $90.08\%$ on NWPU-RESISC45 under the two training ratios, respectively. Although these results are still lower than full-image visual references, they demonstrate that textual semantic proxies can support meaningful scene recognition without using image features during classification.

The performance gap between text-only and vision-based recognition varies across datasets and should be interpreted together with dataset complexity and the saturation level of visual baselines. On AID, G3+Flat reaches $93.95\%$ and $95.18\%$, approaching some vision-based references, which shows that semantic proxies can preserve substantial high-level land-use information. On UCM, SeProRS also achieves high accuracy, but visual methods are nearly saturated, mostly above $98\%$, making the remaining gap more visible despite the strong text-only performance. On NWPU-RESISC45, the gap is larger, suggesting that textual proxies are less able to preserve fine-grained visual cues, spatial layouts, and subtle category differences when the dataset contains more classes and stronger intra-class diversity.

\subsection{Effect of VLMs}
The performance of SeProRS is clearly affected by the proxy generator, yet it does not increase monotonically with model scale. While G3 generally provides the strongest results, G1 outperforms G2 in all matched configurations, indicating that larger VLMs do not always generate more discriminative proxies for text-only RSSC. This result suggests that proxy quality should be evaluated by its downstream semantic utility, including faithfulness, scene specificity, and compatibility with the text embedding space, rather than by generator size alone.

\subsection{Effect of Prompting Strategies}
The Flat Prompt consistently outperforms the Structured Prompt across all VLMs and datasets. This result suggests that a more explicitly organized description does not necessarily lead to better text-only classification. A possible explanation is that flat descriptions tend to produce more coherent scene-level semantics, which better match the representation space of general-purpose text embedding models. In contrast, structured descriptions may introduce rigid formatting or uneven information across semantic fields, which can weaken the overall discriminability of the resulting text representation.

\subsection{Text Embedding Behavior}
\begin{figure}[t]
\centering
{\includegraphics[width=\linewidth]{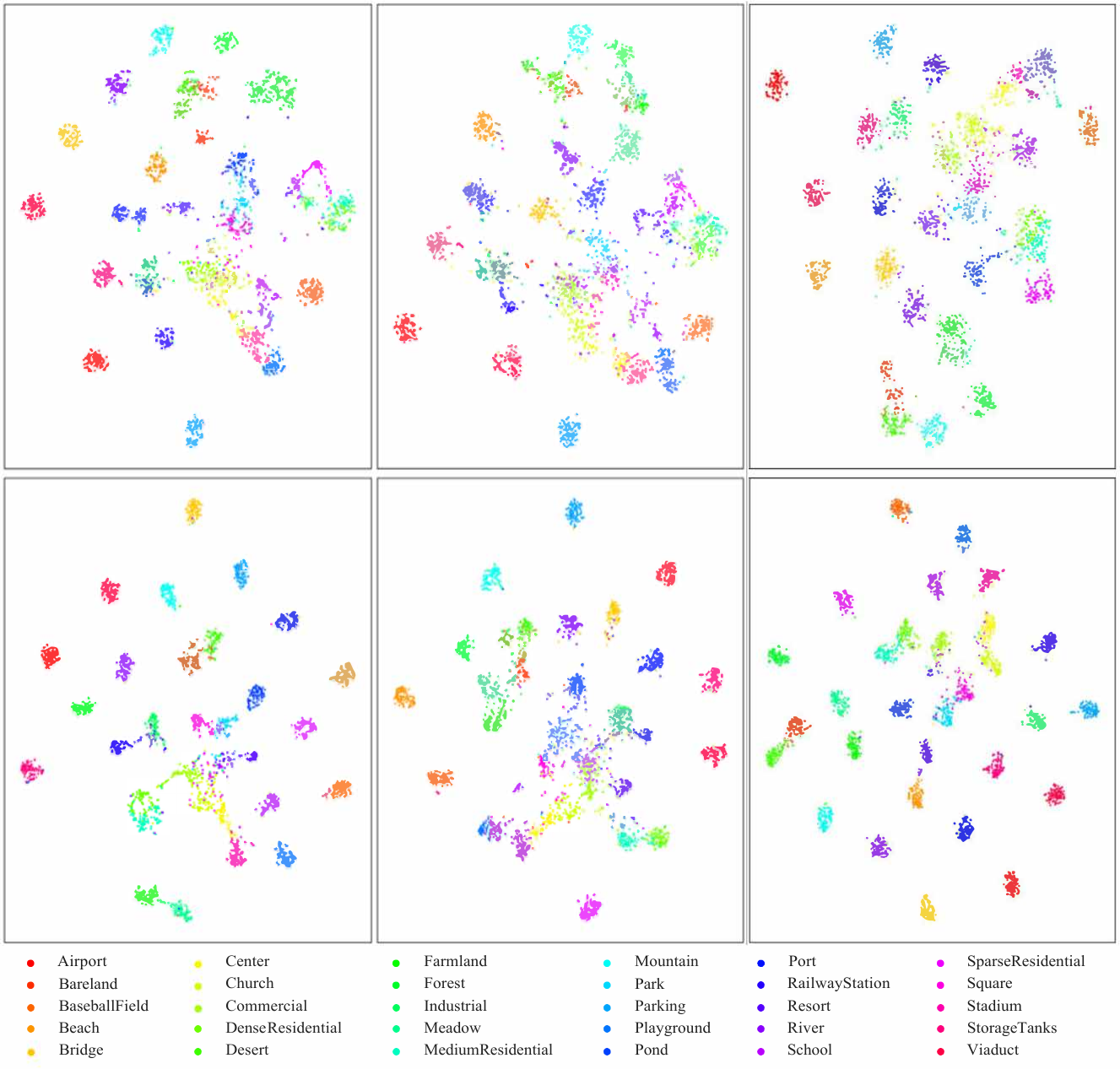}}
\caption{t-SNE visualization of semantic proxy features on AID under the 50\% training setting with Flat Prompt. Columns correspond to G1, G2, and G3, respectively. The first and second rows show the text embeddings before and after the lightweight projection layer, which serves as a text-feature adapter for RSSC.}
\label{fig_tsne}
\end{figure}

Fig.~\ref{fig_tsne} visualizes the text embeddings of semantic proxies using t-SNE on AID. First, many categories form recognizable clusters, indicating that the generated proxies preserve class-discriminative semantic information rather than acting as generic descriptions. Second, the visualization further shows that proxy generation quality is reflected in the text feature space. The separability of different generators is broadly consistent with the quantitative trend in Tables~\ref{tab:ucm_image_reference}-\ref{tab:nwpu_image_reference}, where G3 exhibits the clearest class structures, followed by G1, while G2 shows more noticeable overlaps. Finally, the difference between the two rows demonstrates that the lightweight projection layer makes the class structures more compact and separable, indicating its role as a simple text-feature adapter for aligning general-purpose text embeddings with the RSSC label space.

\subsection{Cross-Generator Transfer}
\begin{table}[t]\small
\centering
\setlength{\tabcolsep}{6pt}
\renewcommand{\arraystretch}{1.08}
\caption{Cross-generator transfer accuracy on AID with the 50\% training split and Flat Prompt.}
\label{tab:cross_vlm_transfer}
\begin{tabular}{cccc}
\toprule
\textbf{Train/Test} & \textbf{G1} & \textbf{G2} & \textbf{G3} \\
\midrule
\textbf{G1} & $\mathbf{92.08}$ & $79.66$ & $80.56$ \\
\textbf{G2} & $82.16$ & $\mathbf{88.54}$ & $82.10$ \\
\textbf{G3} & $78.78$ & $78.40$ & $\mathbf{95.16}$ \\
\bottomrule
\end{tabular}
\end{table}

Table~\ref{tab:cross_vlm_transfer} presents the cross-generator transfer results on AID under the 50\% training setting with Flat Prompt. The diagonal entries correspond to the standard same-generator setting, while the off-diagonal entries evaluate whether a classifier trained on proxies from one VLM can generalize to proxies generated by another VLM. A clear observation is that the diagonal results are consistently higher than most off-diagonal results, indicating that semantic proxies exhibit generator-specific distributions. This suggests that different VLMs may describe the same remote sensing scene with different language styles, semantic granularity, or descriptive preferences.

Interestingly, G3 achieves the highest same-generator accuracy, reaching $95.16\%$, but its proxies transfer poorly to G1 and G2, with accuracies of $78.78\%$ and $78.40\%$, respectively. This indicates that a stronger proxy generator can produce highly discriminative proxies within its own semantic space, but these proxies may also introduce a more generator-specific representation style. In contrast, G2 shows more balanced cross-generator behavior: although its same-generator accuracy is lower than G3, training on G2 proxies transfers better to both G1 and G3. These results reveal that proxy quality is not determined solely by same-generator classification accuracy; transferability across VLMs is another important property of semantic proxies.

Overall, the off-diagonal results remain substantially above chance, showing that different VLMs still preserve shared scene-level semantic evidence. However, the gap between diagonal and off-diagonal performance reveals a proxy distribution shift across generators. This finding suggests that future text-only RSSC methods may benefit from generator-invariant proxy learning or proxy normalization strategies, especially when semantic proxies are collected from heterogeneous VLM sources.

\subsection{Efficiency and Scalability}
\begin{table*}[t]\small
\centering
\caption{Efficiency comparison on AID under the 50\%-50\% setting.}
\label{tab:efficiency}
\begin{tabular}{llccccc}
\toprule
\textbf{Type} & \textbf{Method} & \textbf{Max OA (\%)} & \textbf{Params (M)} & \textbf{Train Time (s/it)} & \textbf{Test Time (s/it)} & \textbf{Max Mem. (MB)} \\
\midrule
Vision & DINOv3-ViT-B/16 Adapter & 98.58 & 7.12 & 0.3025 & 0.2861 & 8543 \\
Vision & UniRepLKNet-T & 97.88 & 31.07 & 0.4557 & 0.3030 & 15793 \\
\midrule
Text & SeProRS & 95.20 & 0.54 & 0.0701 & 0.0935 & 1266 \\
\bottomrule
\end{tabular}
\end{table*}

Table~\ref{tab:efficiency} reports the efficiency comparison on AID under the 50\%-50\% setting. SeProRS uses only $0.54$M trainable parameters, much fewer than DINOv3-Adapter ($7.12$M) and UniRepLKNet-T ($31.07$M). It also achieves substantially lower peak memory consumption, requiring only $1266$ MB compared with $8543$ MB and $15793$ MB for the two vision-based methods. In terms of speed, SeProRS needs $0.0701$ s/it for training and $0.0935$ s/it for testing, making it clearly faster than the visual references.

These results show that, after semantic proxies are generated, the downstream text-only classifier is lightweight and scalable. Although the offline VLM-based proxy generation cost is not included, the generated proxies can be reused across different classifiers or analyses. Thus, SeProRS provides an efficient baseline for modality-reduced RSSC, trading a moderate accuracy gap for significantly lower trainable parameters, memory usage, and computational cost.

\subsection{Limitations}
The experimental results clarify the potential of text-only RSSC while also exposing the limitations introduced by semantic abstraction.

First, textual semantic proxies can serve as partial substitutes for remote sensing images. Although SeProRS does not access image pixels or visual features during classification, it achieves meaningful performance across three standard datasets. This indicates that high-level scene evidence, such as dominant objects, land-use functions, spatial context, and object co-occurrence, can be transferred from images to text. However, this substitution is incomplete, since fine-grained visual cues such as texture, scale, geometry, and subtle spatial layouts are difficult to fully preserve in natural language.

Second, the cross-generator transfer results suggest the potential of more flexible image-free inference. A classifier trained on proxies generated by one VLM can still obtain non-trivial performance on proxies generated by another VLM, indicating that different generators preserve shared scene-level semantics. This means that, once a semantic proxy is available, the final classification stage does not depend on the original image or a fixed VLM generator. Meanwhile, the performance drop in cross-generator settings reveals generator-specific proxy shifts, suggesting that generator-invariant proxy learning is a promising future direction.

Third, SeProRS is intended as a basic and reproducible baseline rather than a fully optimized solution. Its components are deliberately simple, including prompt-based proxy generation, general-purpose text encoding, a lightweight projection adapter, and a linear classifier. This design helps isolate the role of semantic proxies, but also leaves considerable room for improvement. Future work may further narrow the gap to vision-based methods by improving proxy generation, designing remote sensing-oriented text encoders, refining projection adapters, or introducing proxy quality estimation.

\section{Conclusion}
This paper introduced text-only RSSC, a new setting that studies whether remote sensing scenes can be classified from textual semantic proxies alone. We proposed SeProRS as a reproducible semantic proxy baseline and established a benchmark using three datasets, three VLMs, and two prompting strategies. Experiments show that semantic proxies can preserve substantial discriminative information, while also exposing limitations caused by information loss and proxy generation quality. Additional implementation details and proxy examples further indicate that the benchmark is reproducible and inspectable: the VLMs generate label-free textual proxies, and the downstream classifier operates entirely in the textual space. We hope this work will stimulate future research on semantic proxy-based and modality-reduced remote sensing understanding.

\small
\bibliographystyle{IEEEtran}
\bibliography{references}

\clearpage
\normalsize
\appendices
\section{Additional Implementation Details}
\label{app:implementation}
This appendix reports the classifier training settings and complete representative semantic proxies. Table~\ref{tab:text_hyperparams} summarizes the text-only classifier configuration, and Table~\ref{tab:visual_hyperparams} lists the settings of the vision-based reference models. The semantic proxy generation protocol is reported in Table~\ref{tab:proxy_generation_hyperparams} of the main text.

The training implementation scales the base learning rate linearly with the global batch size, i.e., $\eta_{\mathrm{eff}}=\eta_{\mathrm{base}}B_{\mathrm{global}}/512$. With $\eta_{\mathrm{base}}=5\times10^{-4}$ and $B_{\mathrm{global}}=128$, the effective learning rate is $1.25\times10^{-4}$ for both text-only and vision-based training, as confirmed by the training logs. The scheduler uses five warm-up epochs starting from $1\times10^{-6}$, followed by cosine decay with a minimum learning rate of $1\times10^{-6}$.

\begin{table}[t]\small
\centering
\caption{Hyperparameters for text-only classification.}
\label{tab:text_hyperparams}
\resizebox{\columnwidth}{!}{
\begin{tabular}{ll}
\toprule
\textbf{Item} & \textbf{Setting} \\
\midrule
Text encoder & \emph{bge-base-en-v1.5} \\
Encoder update & Frozen \\
Maximum text length & 192 tokens \\
Projection adapter & LayerNorm, Linear $768{\rightarrow}512$, GELU, Dropout \\
Classifier head & LayerNorm, Linear $512{\rightarrow}256$, GELU, Dropout, Linear \\
Loss & Cross-entropy \\
Optimizer & AdamW \\
Base learning rate & $5\times10^{-4}$ \\
Effective learning rate & $1.25\times10^{-4}$ \\
Warm-up learning rate & $1\times10^{-6}$ \\
Minimum learning rate & $1\times10^{-6}$ \\
Weight decay & 0.05 \\
Schedule & Cosine schedule with 5 warm-up epochs \\
Epochs & 200 \\
Batch size & 128 \\
Runs & 5 independent seeds \\
\bottomrule
\end{tabular}
}
\end{table}

\begin{table}[t]\small
\centering
\caption{Hyperparameters for vision-based reference methods.}
\label{tab:visual_hyperparams}
\setlength{\tabcolsep}{5pt}
\begin{tabular}{lcc}
\toprule
\textbf{Item} & \textbf{DINOv3 Adapter} & \textbf{UniRepLKNet-T} \\
\midrule
Input size & $224\times224$ & $224\times224$ \\
Pretraining & DINOv3 ViT-B/16 & ImageNet \\
Training epochs & 200 & 200 \\
Batch size & 128 & 128 \\
Optimizer & AdamW & AdamW \\
Base learning rate & $5\times10^{-4}$ & $5\times10^{-4}$ \\
Effective learning rate & $1.25\times10^{-4}$ & $1.25\times10^{-4}$ \\
Warm-up learning rate & $1\times10^{-6}$ & $1\times10^{-6}$ \\
Minimum learning rate & $1\times10^{-6}$ & $1\times10^{-6}$ \\
Weight decay & 0.05 & 0.05 \\
Schedule & Cosine & Cosine \\
Warm-up epochs & 5 & 5 \\
\bottomrule
\end{tabular}
\end{table}

\section{Representative Semantic Proxy Examples}
\label{app:proxies}
Tables~\ref{tab:proxy_examples}--\ref{tab:proxy_examples_nwpu} provide complete representative proxies for one image from each of the three datasets. For each image, both Flat and Structured outputs are shown for all three generators: $G_1$ (LLaVA-OneVision-Qwen2-0.5B), $G_2$ (Qwen2.5-VL-3B), and $G_3$ (Gemini-2.5-Flash). Dataset and image identifiers are retained to support comparison across generators and prompting strategies.

These are the original generated descriptions, without manual corrections or abridgment. The full stored proxies are shown; downstream encoding still uses the 192-token limit reported in Table~\ref{tab:text_hyperparams}. Any image identifiers or class names shown for reader interpretation are not included in the generation prompts. The examples illustrate differences in semantic detail and factual reliability; for instance, the $G_2$ Flat proxy for airplane\_001.jpg describes aircraft as being in flight, whereas the other descriptions characterize them as parked. Such generator-dependent statements are retained to document the inputs available to the text-only classifier, rather than presented as verified scene annotations.

\begin{table*}[!t]\scriptsize
\centering
\caption{Representative semantic proxies on UCM for the same image under Flat and Structured prompts. $G_1$, $G_2$, and $G_3$ denote LLaVA-OneVision-Qwen2-0.5B, Qwen2.5-VL-3B, and Gemini-2.5-Flash, respectively.}
\label{tab:proxy_examples}
\setlength{\tabcolsep}{3pt}
\renewcommand{\arraystretch}{1.05}
\begin{tabular}{@{}p{0.10\linewidth}p{0.12\linewidth}p{0.04\linewidth}p{0.09\linewidth}p{\dimexpr0.65\linewidth-24pt\relax}@{}}
\toprule
\textbf{Dataset} & \textbf{Image} & \textbf{VLM} & \textbf{Prompt} & \textbf{Semantic proxy} \\
\midrule
UCM & agricultural00.tif & $G_1$ & Flat & The image depicts a large, well-organized agricultural field with rows of crops. The field is divided into multiple sections by straight lines that run parallel to each other, indicating the use of precision farming techniques. The crops appear to be in good health and are likely being cultivated for their potential yield or market value. There are no visible structures or people within the field, suggesting it may be an open space used for crop cultivation rather than a farmstead. \\
\addlinespace[1pt]
UCM & agricultural00.tif & $G_1$ & Structured & The image shows a large, well-maintained agricultural field with neatly arranged rows of crops. The field is bordered by a dirt road and there are trees visible in the background. Buildings can be seen lining the perimeter of the field, indicating residential or commercial structures. Roads run parallel to the fields, providing access for vehicles and farming equipment. Vegetation includes grasses and other plants that cover the soil, contributing to the overall health of the crop. The scene appears to be an active agricultural area, likely used for growing crops such as grains or vegetables. The presence of roads suggests easy transportation for farm equipment and supplies. The buildings could be storage facilities or offices related to the operations of the farm. \\
\addlinespace[1pt]
UCM & agricultural00.tif & $G_2$ & Flat & The remote sensing image depicts a large agricultural field with rows of crops, likely wheat or corn, stretching across the landscape. The field is bordered by a dense forest on the left side, indicating a transition between agricultural land and natural vegetation. The functional context suggests this area is used for crop cultivation, possibly for commercial purposes such as grain production. \\
\addlinespace[1pt]
UCM & agricultural00.tif & $G_2$ & Structured & The image depicts a large agricultural field with rows of crops, likely wheat or corn, stretching across the landscape. The field is surrounded by a mix of trees and open areas, indicating a rural setting. The functional use of this area is primarily agricultural, designed for crop cultivation and harvesting. \\
\addlinespace[1pt]
UCM & agricultural00.tif & $G_3$ & Flat & An expansive agricultural scene characterized by numerous, precisely aligned green crop rows covering the majority of the area, hinting at large-scale cultivation or a nursery, with a small patch of denser, irregular vegetation in the upper left corner and a bright, rectilinear artificial structure at the bottom right, collectively depicting an organized rural or peri-urban agricultural landscape. \\
\addlinespace[1pt]
UCM & agricultural00.tif & $G_3$ & Structured & The scene depicts a predominantly agricultural landscape characterized by extensive cultivation. It features numerous parallel crop rows, a small patch of natural vegetation, and a potential artificial structure at the periphery. This area is primarily used for agricultural production, indicating a farming context. \\
\bottomrule
\end{tabular}
\end{table*}

\begin{table*}[!t]\scriptsize
\centering
\caption{Representative semantic proxies on AID for the same image under Flat and Structured prompts. $G_1$, $G_2$, and $G_3$ denote LLaVA-OneVision-Qwen2-0.5B, Qwen2.5-VL-3B, and Gemini-2.5-Flash, respectively.}
\label{tab:proxy_examples_aid}
\setlength{\tabcolsep}{3pt}
\renewcommand{\arraystretch}{1.05}
\begin{tabular}{@{}p{0.10\linewidth}p{0.12\linewidth}p{0.04\linewidth}p{0.09\linewidth}p{\dimexpr0.65\linewidth-24pt\relax}@{}}
\toprule
\textbf{Dataset} & \textbf{Image} & \textbf{VLM} & \textbf{Prompt} & \textbf{Semantic proxy} \\
\midrule
AID & airport\_1.jpg & $G_1$ & Flat & The image is an aerial view of a busy airport tarmac, showcasing multiple airplanes parked at gates. The tarmac is marked with various lines and symbols indicating taxiways and other operational areas. In the center of the image, there's a large jet bridge connecting one terminal to another, facilitating passenger boarding and deplaning. Adjacent to this jet bridge, several smaller aircraft are stationed on the ground, likely awaiting their turn for departure or having just arrived. The overall scene represents a typical day at an airport, with planes ready for takeoff and passengers waiting in anticipation. \\
\addlinespace[1pt]
AID & airport\_1.jpg & $G_1$ & Structured & The image shows an aerial view of a busy airport terminal with multiple jet bridges connecting the terminals. There are several airplanes parked at gates, and various buildings and roads can be seen surrounding the airport. Buildings include large hangars, terminals, and control towers, while roads are visible as well as parking areas for vehicles. Vegetation is present in the form of grassy patches near the terminal buildings. The scene represents transportation-related activities, with airplanes being the primary mode of transport. The presence of jet bridges indicates that passengers can board or disembark from aircraft directly to the terminal. The overall setting suggests a commercial airport environment. \\
\addlinespace[1pt]
AID & airport\_1.jpg & $G_2$ & Flat & The image depicts an expansive airport terminal with multiple runways and taxiways, featuring several aircraft parked at gates. The terminal is surrounded by various ground support equipment and vehicles, indicating active operations. The surrounding area includes parking lots and possibly maintenance facilities, suggesting a well-equipped airport infrastructure. \\
\addlinespace[1pt]
AID & airport\_1.jpg & $G_2$ & Structured & The image depicts an expansive airport terminal with multiple runways and taxiways, surrounded by various infrastructure elements including parking lots and service areas. The terminal is characterized by its long, linear structure with numerous windows and doors, indicative of passenger boarding and disembarking points. The functional aspect of this scene is primarily transportation, serving as a hub for air travel with clear demarcations for aircraft movement and ground operations. \\
\addlinespace[1pt]
AID & airport\_1.jpg & $G_3$ & Flat & An expansive airport facility is depicted, featuring a prominent, elongated passenger terminal building with numerous aircraft parked at gates along its length, surrounded by vast concrete aprons, intricate taxiways, and portions of runways, alongside a distinct landscaped circular structure near the main terminal, all indicative of a busy aviation hub designed for air travel and transport. \\
\addlinespace[1pt]
AID & airport\_1.jpg & $G_3$ & Structured & The scene depicts a large-scale developed area characterized by extensive paved surfaces and structures, indicative of an airport. Numerous aircraft are visible parked alongside a long, multi-sectioned terminal building, surrounded by an intricate network of runways, taxiways, and aprons. This area serves as a critical hub for air transportation, facilitating the movement and servicing of aircraft and passengers. \\
\bottomrule
\end{tabular}
\end{table*}

\begin{table*}[!t]\scriptsize
\centering
\caption{Representative semantic proxies on NWPU-RESISC45 for the same image under Flat and Structured prompts. $G_1$, $G_2$, and $G_3$ denote LLaVA-OneVision-Qwen2-0.5B, Qwen2.5-VL-3B, and Gemini-2.5-Flash, respectively.}
\label{tab:proxy_examples_nwpu}
\setlength{\tabcolsep}{3pt}
\renewcommand{\arraystretch}{1.05}
\begin{tabular}{@{}p{0.10\linewidth}p{0.12\linewidth}p{0.04\linewidth}p{0.09\linewidth}p{\dimexpr0.65\linewidth-24pt\relax}@{}}
\toprule
\textbf{Dataset} & \textbf{Image} & \textbf{VLM} & \textbf{Prompt} & \textbf{Semantic proxy} \\
\midrule
NWPU-RESISC45 & airplane\_001.jpg & $G_1$ & Flat & The image shows a scene of three airplanes parked on an airport tarmac. The planes are positioned in a staggered formation, with the middle plane slightly ahead of the other two. Each airplane is white and appears to be a commercial jet airliner. The tarmac surface is a light grey color, typical for such environments. There are no visible markings or identifiers that would indicate the specific location or airline. The overall scene type is remote sensing, as it captures aerial imagery rather than ground-level observations. \\
\addlinespace[1pt]
NWPU-RESISC45 & airplane\_001.jpg & $G_1$ & Structured & The image shows a large, open area with three airplanes parked on the tarmac. The planes are white and appear to be commercial airliners. There is also some vegetation visible in the background. The airplanes are likely used for commercial purposes such as air travel or cargo transport. The tarmac appears to be made of concrete and there are no other objects or structures visible in the immediate vicinity. The scene suggests that it could be an airport or a transportation hub where aircraft are parked before takeoff or after landing. The presence of multiple airplanes indicates that this location serves a significant number of passengers or cargo operations. \\
\addlinespace[1pt]
NWPU-RESISC45 & airplane\_001.jpg & $G_2$ & Flat & The image depicts an aerial view of an airport runway with three airplanes in flight. Two aircraft are positioned closer to the camera, appearing larger due to their proximity, while the third is further away, smaller in size. The planes are likely commercial jets, given their size and design, and they are flying at a low altitude, suggesting they might be preparing for landing or taking off. The runway is clearly marked, indicating its functional use for aircraft operations. \\
\addlinespace[1pt]
NWPU-RESISC45 & airplane\_001.jpg & $G_2$ & Structured & The image depicts an airport with multiple airplanes on the tarmac. The ground is paved with asphalt, indicating a well-maintained runway area. Airplanes are present, suggesting active air traffic at this location. \\
\addlinespace[1pt]
NWPU-RESISC45 & airplane\_001.jpg & $G_3$ & Flat & An airbase or large airport scene is presented, characterized by multiple large, propeller-driven transport aircraft prominently parked on a paved tarmac or apron, strongly indicating a military or heavy logistics operational facility. \\
\addlinespace[1pt]
NWPU-RESISC45 & airplane\_001.jpg & $G_3$ & Structured & This is an aerial scene dominated by an airport or airfield environment. Multiple propeller aircraft are visible parked on large, untextured concrete or asphalt surfaces. The scene clearly indicates a transportation function, specifically for air travel and aircraft operations. \\
\bottomrule
\end{tabular}
\end{table*}

\end{document}